\documentclass{article}
\usepackage{ijcai24}

\usepackage{times}
\usepackage{soul}
\usepackage{url}
\usepackage[hidelinks]{hyperref}
\usepackage[utf8]{inputenc}
\usepackage[small]{caption}
\usepackage{graphicx}
\usepackage{amsmath}
\usepackage{amsthm}
\usepackage{booktabs}
\usepackage{algorithm}
\usepackage{algpseudocode}
\usepackage[switch]{lineno}
\usepackage{multirow}
\usepackage{subcaption}
\usepackage{xcolor}
\usepackage{makecell}

\usepackage{tikz}
\newcommand*\circled[1]{\tikz[baseline=(char.base)]{
            \node[shape=circle,draw,inner sep=0.1pt] (char) {#1};}}

\linenumbers

\newcommand{\systemName}{LLMem}

\title{\systemName: Estimating GPU Memory Usage for Fine-Tuning Pre-Trained LLMs}

\author{Taeho Kim$^1$\footnote{This work is based on Taeho Kim's internship at AWS.} \and Yanming Wang$^2$ \and Vatshank Chaturvedi$^2$ \and Lokesh Gupta$^2$ \and Seyeon Kim$^1$ \and Yongin Kwon$^3$ \and Sangtae Ha$^1$
\affiliations
$^1$University of Colorado Boulder\\
$^2$Amazon Web Services\footnote{This work is unconnected to current role at AWS.}\\
$^3$Electronics and Telecommunications Research Institute
\emails
\{taeho.kim,seyeon.kim,sangtae.ha\}@colorado.edu,
\{yanmwang,vatshc,lokeshgu\}@amazon.com,
yongin.kwon@etri.re.kr
}

\begin{document}
\nolinenumbers

\maketitle

\begin{abstract}
Fine-tuning pre-trained large language models (LLMs) with limited hardware presents challenges due to GPU memory constraints. Various distributed fine-tuning methods have been proposed to alleviate memory constraints on GPU. However, determining the most effective method for achieving rapid fine-tuning while preventing GPU out-of-memory issues in a given environment remains unclear. To address this challenge, we introduce \systemName, a solution that estimates the GPU memory consumption when applying distributed fine-tuning methods across multiple GPUs and identifies the optimal method. We conduct GPU memory usage estimation prior to fine-tuning, leveraging the fundamental structure of transformer-based decoder models and the memory usage distribution of each method. Experimental results show that \systemName\ accurately estimates peak GPU memory usage on a single GPU, with error rates of up to 1.6\%. Additionally, it shows an average error rate of 3.0\% when applying distributed fine-tuning methods to LLMs with more than a billion parameters on multi-GPU setups.
\end{abstract}
\section{Introduction} \label{section1}

Since the introduction of the Transformer model~\cite{vaswani2017attention}, researchers have proposed numerous language models based on it. As the model's performance has improved, its size has grown exponentially, necessitating a substantial dataset for training. However, training emerging large language models (LLMs) is infeasible without a dedicated infrastructure with high-performance hardware due to memory constraints. Instead, it is preferred to utilize a small dataset to fine-tune a pre-trained model for a specific application.

To efficiently handle small datasets and reduce training time, the conventional method of data parallelism places the entire model on each GPU, splits the dataset, and trains simultaneously. Nevertheless, the model size remains huge, potentially causing GPU out-of-memory (OOM) issues. Therefore, it is necessary to reduce the amount of memory a GPU uses by splitting the model and distributing it to each GPU.

ZeRO~\cite{rajbhandari2020zero} Stage 3 is an advanced data parallelism method that partitions the model parameters, gradients, and optimizer states to each GPU for memory advantage while maintaining the distribution of the dataset across GPUs. Although ZeRO Stage 3 saves memory by using only partitioned model data on each GPU during non-computation phases, there are limitations in preventing GPU OOM issues because partitioned parameters/gradients must be all-gathered during computation.

Tensor parallelism divides each parameter tensor in the model into rows or columns and distributes them to each GPU, using only partitioned parameters on each GPU during computation. For example, Megatron-LM~\cite{shoeybi2019megatron}, a representative tensor parallelism method, splits a tensor along its rows or columns considering the position and connection of operators. By doing so, it can reduce GPU memory usage more than data parallelism when the model size is large.

As we described above, various distributed fine-tuning methods have been proposed, but the GPU memory usage and fine-tuning time required for each are different. For instance, conventional data parallelism provides the shortest fine-tuning time but requires the highest GPU memory usage. On the other hand, tensor parallelism has no benefit in saving fine-tuning time but can significantly reduce GPU memory usage. Users may want to select an appropriate method that avoids GPU OOM and has a short fine-tuning time. However, it is difficult to determine in advance whether there is enough GPU memory to fine-tune a given pre-trained LLM.



DNNMem~\cite{gao2020estimating} is the most recent work detailing procedures to estimate GPU memory usage on a single GPU. DNNMem provides key equations for GPU memory estimation when training various DNN models by analyzing the connections between operators and live tensors in the forward and backward passes. However, it has limitations for fine-tuning LLMs. GPU memory estimation for fine-tuning transformer-based LLM is challenging for two reasons. 

First, when fine-tuning an LLM in multi-GPU, distributed fine-tuning methods should be used to overcome GPU memory constraints due to large model sizes. Depending on the method used, the distribution of parameters, gradients, and optimizer states to each GPU is different, as is the amount of GPU memory used during the calculation process. Therefore, GPU memory usage estimates from a single GPU cannot be used in a multi-GPU environment. 

Second, GPU memory consumption must be predicted by distinguishing between transformer and language modeling head (lm\_head) parts. The transformer part is the central part of fine-tuning, where chunk memory management for memory sharing of model parameters and gradients is applied, and parameters are updated. On the other hand, the lm\_head part requires separate analysis because it does not apply distributed methods directly and consumes a lot of memory due to its large dictionary size. 

To address these challenges, we propose \systemName\ that estimates the GPU memory consumption when applying distributed fine-tuning methods to multiple GPUs. 
\systemName\ considers several factors to estimate GPU memory usage for each method, including recombining parameters prior to computation when applying advanced data parallelism and the output driven by all-gather in the backward pass when using tensor parallelism. Additionally, \systemName\ analyzes the difference in memory allocation method between the transformer and the lm\_head part and reflects it in GPU memory estimation.
To the best of our knowledge, this is the first work to estimate the peak GPU memory consumption for LLM fine-tuning.

In summary, our contributions are:

\begin{itemize}
    \item We propose a GPU memory usage estimation method for LLM fine-tuning on single and multiple GPUs.
    \item We provide an algorithm to determine the most efficient distributed fine-tuning method based on GPU memory usage estimation.
    \item Experimental results show that \systemName\ estimates peak GPU memory usage to fine-tune LLM on a single GPU with error rates of up to 1.6\%, which is significantly smaller than the state-of-the-art DNNMem's average error rate of 42.6\%. When applying distributed fine-tuning methods to LLMs with over a billion parameters on multiple GPUs, \systemName\ successfully estimates GPU memory usage with an average error rate of 3.0\%. 
\end{itemize}


Our source code repository can be found at~\href{https://github.com/taehokim20/LLMem}{https://github.com/taehokim20/LLMem}.
\section{Related Works}

\subsection{GPU Memory Estimation}
There have been several attempts to avoid GPU OOM issues by predicting the GPU memory usage that will be used to train a given model in advance. DNNMem~\cite{gao2020estimating} sequentially traverses the computation graph of a DL model and computes the GPU memory consumption by taking into account previously allocated but still in-use tensors, newly allocated tensors for the currently visited operator, and resident buffers of the CUDA context and allocator reservation. Our \systemName\ is inspired by DNNMem, whose mechanism is described in more detail in Section~\ref{section3}. TSplit~\cite{nie2022tsplit} also calculates the total size of live tensors for the visiting operator. However, TSplit lacks an explanation of the detailed memory estimation process and its accuracy. SchedTune~\cite{albahar2022schedtune} predicts GPU memory usage not only based on DL model characteristics but also on different GPU types running the job. 
However, using measured GPU memory as data for prediction does not align with the purpose of estimating memory usage before fine-tuning a model.

\subsection{Distributed Fine-Tuning with GPUs}
Data parallelism can enhance fine-tuning speed in proportion to the number of GPUs. However, LLM often runs into memory constraints, so the ZeRO optimizer~\cite{rajbhandari2020zero}, described in Section~\ref{section1}, is widely used as an alternative.
The ZeRO optimizer selectively gathers only the model parameters or gradients required during the computation process and utilizes reduce-scatter after the computation to maintain their partitioning on each GPU.

Tensor parallelism can further reduce peak GPU memory usage by sharding tensors under certain conditions, eliminating the need to gather all model parameters and gradients even during computation. Tensor parallelism results in each GPU producing only partial results, necessitating that all GPUs receive the same input data. 
The widely adopted tensor parallelism method, Megatron-LM~\cite{shoeybi2019megatron}, splits each model parameter tensor by row or column. Other proposed methods~\cite{xu2021efficient}~\cite{wang20212}~\cite{bian2021maximizing} achieve additional memory savings by sharding both input and model parameters.


If GPU memory constraints cannot be met with any distributed fine-tuning method on GPUs alone, we can use heterogeneous fine-tuning utilizing CPU memory. ZeRO-offload~\cite{ren2021zero} manages gradients, optimizer states, and optimizer computation on the CPU while retaining parameters and forward and backward computation on the GPU. 

\section{Motivation} \label{section3}

To select distributed fine-tuning methods, it is crucial to estimate GPU memory usage accurately. Existing approaches for estimating GPU memory usage do not consider scenarios where advanced data parallelism, such as the ZeRO Stage 3 optimizer~\cite{rajbhandari2020zero}, or tensor parallelism is applied across multiple GPUs. 
Relying solely on estimated GPU memory usage on a single GPU when estimating on multiple GPUs can lead to significant errors. 
In this section, we implement the existing DNNMem~\cite{gao2020estimating}, validate the implementation results, and discuss factors causing substantial GPU memory estimation errors during the fine-tuning of pre-trained transformer-based language models.

\subsection{DNNMem Implementation} \label{section3-1}
DNNMem source codes are not publicly available and are mainly based on TensorFlow, so we implement DNNMem based on the description in the paper~\cite{gao2020estimating}. First, we extract the corresponding computation graph from a given pre-trained DL model to identify the output size in each operator based on parameters, batch size ($bs$), and sequence length ($sl$). We also compute pre-allocated GPU memory, including CUDA context and weight tensors of the model, before operator execution. In particular, since PyTorch does not release the loaded model parameters until the end of the fine-tuning, the initial GPU memory is retained throughout the fine-tuning process. The next step is to compute peak GPU memory usage at each operator while traversing the graph. We compute additional GPU memory with the input/output tensors and previously unreleased tensors in each operator during the forward propagation. Additionally, we reflect that PyTorch aligns with multiples of 512 bytes for internal tensor fragmentation, and DNNMem treats the buffer size as a constant (64 MB by default) as memory block management. 

To validate our DNNMem implementation, we compare GPU memory estimation results for the BERT~\cite{devlin2018bert} model on the 
GLUE benchmark~\cite{wang2018glue} with the experimental results from the paper. The environment we used in the experiment was PyTorch 2.0.1 with CUDA 11.7 on NVIDIA RTX2060, which differs from PyTorch 1.2.0 with CUDA 9.0 on NVIDIA Tesla P40 used in the DNNMem paper. In $bs=32, sl=32$, our DNNMem shows 34.38\%, and the DNNMem shows 31.42\% error rates. In $bs=64, sl=32$, our DNNMem shows 20.48\%, and the DNNMem shows 19.12\% error rates. Considering similar error rates, we use our DNNMem implementation for single-GPU comparisons.

\begin{figure}[t]
\begin{center}
    \includegraphics[width=0.75\linewidth]{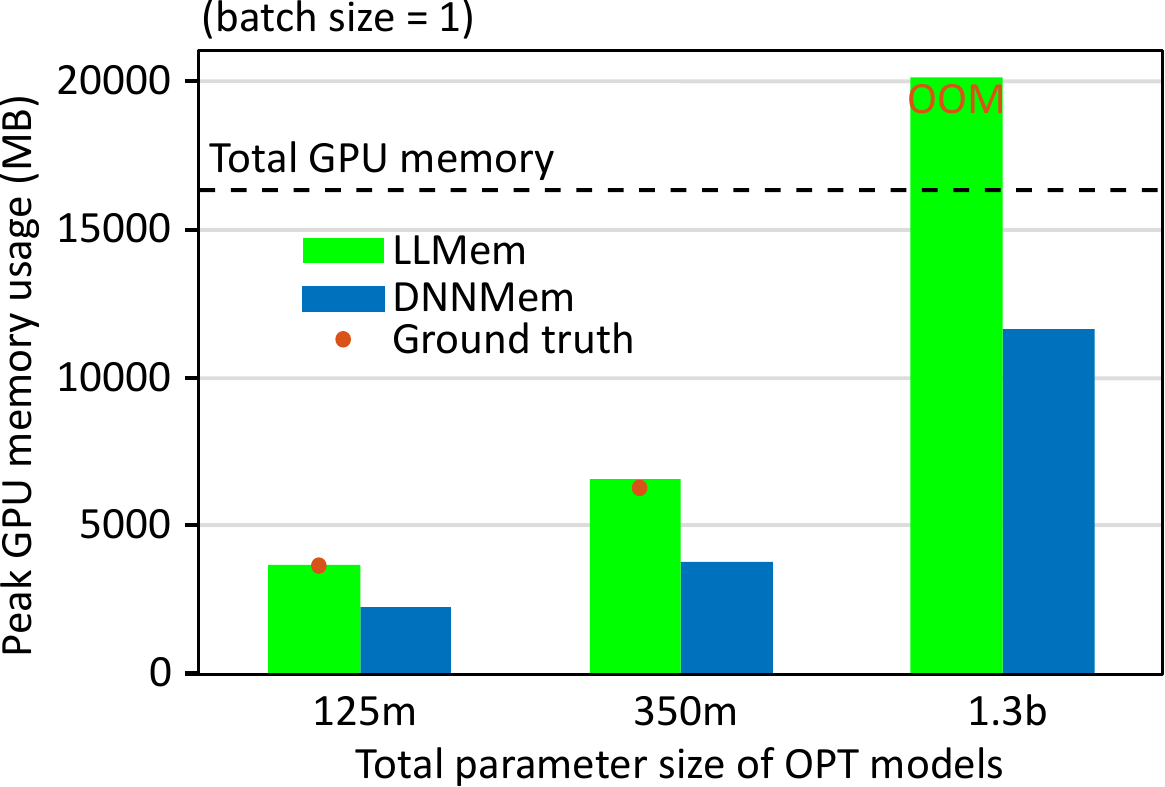}
\end{center}
\caption{Peak GPU memory estimates per total parameter size on a single GPU}
\label{fig:llm-dnn-params}
\end{figure}

\subsection{Limitations of DNNMem for LLM Fine-Tuning Memory Estimation} \label{section3-2}

DNNMem~\cite{gao2020estimating} does not handle mixed precision, which is commonly used in fine-tuning pre-trained language models. In addition, it does not consider how memory chunks are managed to ensure that forward pass parameters and backward pass gradients share the same GPU memory space~\cite{fang2022parallel}. Furthermore, DNNMem overlooks extra GPU memory usage during the initial fine-tuning iteration due to optimizer states. 

Comparison results for estimating peak GPU memory usage of our proposed \systemName\ and DNNMem on a single GPU are shown in Figure~\ref{fig:llm-dnn-params}. The experimental environment is summarized in Section~\ref{setup}. \systemName\ predicts peak GPU memory usage with minimal error rates compared to ground truth, outperforming DNNMem. DNNMem exhibits larger errors as the total parameter size increases. Furthermore, DNNMem fails to predict GPU memory consumption in the context of distributed fine-tuning methods across multiple GPUs. As a result, existing approaches for estimating GPU memory usage face challenges when using the current transformer-based LLM for distributed fine-tuning. 
\begin{table}[t]
\caption{Notation}\label{tab:notation}
\begin{center}
\begin{tabular}{ |p{1.3cm}|p{6.4cm}| } 
\hline
\thead{\textbf{Symbol}} & \thead{\textbf{Description}} \\
\hline
$m_{base}$ & {\small the initially used GPU memory} \\ \hline
$embed_p$ & {\small the input embedding param size} \\ \hline
$lm_p$ & {\small the language modeling head param size} \\ \hline
$cs, bs, sl$ & {\small chunk size, batch size, sequence length} \\ \hline
$other_p$ & {\footnotesize the remaining param size w/o $embed_p$ and $lm_p$} \\ \hline
$B_{16}$, $B_{32}$ & {\small 2 bytes for fp16, 4 bytes for fp32} \\ \hline
$m_p$ & {\footnotesize the GPU memory used by param fp16 and fp32} \\ \hline
$m_{p,16}$ & {\footnotesize the GPU memory used by param fp16} \\ \hline
$m_{p,32}$ & {\footnotesize the GPU memory used by param fp32} \\ \hline
$cu_p$ & {\small the CUDA memory page size} \\ \hline
$m_{os}$ & {\scriptsize the GPU memory used by momentum fp32 and variance fp32} \\ \hline
$e_n, l_n$ & {\footnotesize the number of Embedding or layers} \\ \hline 
$o_n$ & {\small the number of model's output features} \\ \hline
$m_{out}$ & {\small the peak GPU memory usage due to output tensors} \\ \hline
$dict_n$ & {\small the size of the embedding dictionary} \\ \hline
$m_{lm}$ & {\scriptsize the GPU memory used in the lm\_head with the loss calculation} \\ \hline
$m_{peak}^s$ & {\small the peak GPU memory usage on a single GPU} \\ \hline
$gpu_n$ & {\small the number of GPUs in use} \\ \hline
$m_{peak}^{dp}$ & {\small the peak GPU memory usage with the advanced DP on multiple GPUs} \\ \hline
$m_{back}^{tp}$ & {\small the additional GPU memory usage due to the temporary buffer through the backward all-gather} \\ \hline
$dp_n, tp_n$ & {\footnotesize the number of GPUs used for DP or TP} \\ \hline
$m_{peak}^{tp}$ & {\small the peak GPU memory usage with 1D TP on multiple GPUs} \\ \hline
$m_{peak}^{dp+tp}$ & {\small the peak GPU memory usage with the combination of DP+TP on multiple GPUs} \\ \hline
$m_{total}$ & {\small the total GPU memory capacity} \\ \hline
\end{tabular}
\end{center}
\end{table}

\section{Single-GPU Memory Usage Estimation} \label{section4}

This section outlines considerations for estimating GPU memory usage of transformer-based language models on a single GPU. The symbols used in the explanation are organized in Table~\ref{tab:notation}. 

\begin{figure}[t]
\begin{center}
    \includegraphics[width=1\linewidth]{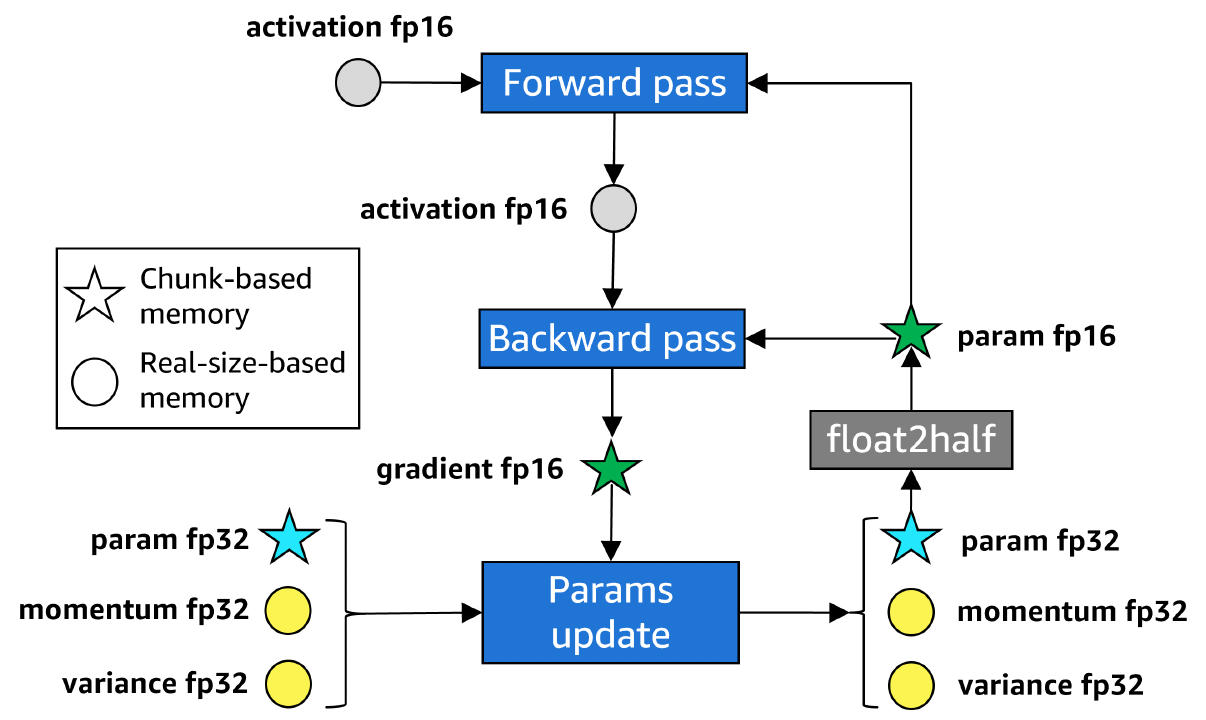}
\end{center}
\caption{Illustration of tensors using GPU memory while fine-tuning the pre-trained model~\protect\cite{ren2021zero}}
\label{fig:tensors}
\end{figure}


\subsection{Workflow for Fine-Tuning Pre-Trained Models} \label{section4-1}

\textbf{Initialization phase.} The initialization phase preceding fine-tuning involves allocating memory for the CUDA context, responsible for managing information to control GPU devices, and memory for applying chunk-based memory management~\cite{fang2022parallel}. The initially used GPU memory is denoted as $m_{base}$. The chunk manager determines the optimal chunk size to minimize GPU memory waste based on the parameters of the provided pre-trained language model. GPU memory spaces for param fp16 (float-16) and param fp32 (float-32) are allocated in units of the chunk size (\circled{1} in Figure~\ref{fig:peak-mems}). 

\noindent\textbf{Fine-tuning phase.} During the fine-tuning phase, param fp16 goes through forward and backward passes, and param fp16 is converted to gradient fp16, as illustrated in Figure~\ref{fig:tensors}. Consequently, param fp16 and gradient fp16 share the same GPU memory space. After the backward pass, the ADAM optimizer updates parameters using optimizer states, including param fp32, momentum fp32, and variance fp32 tensors. Momentum fp32 and variance fp32 tensors, which are not allocated memory during the initialization process before fine-tuning, consume GPU memory based on the actual tensor size, not the chunk size. GPU memory occupied by these tensor types is allocated in the first iteration for fine-tuning (\circled{5} in Figure~\ref{fig:peak-mems}). Subsequently, similar to chunk-based parameters, the GPU memory is retained until the fine-tuning process is complete. 

\begin{figure}[t]
\begin{center}
    \includegraphics[width=0.99\linewidth]{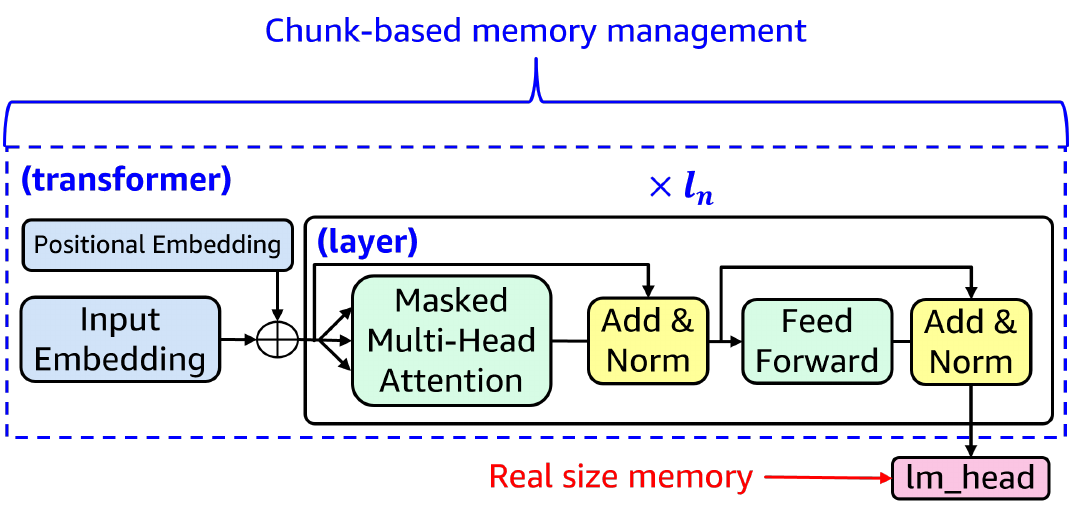}
\end{center}
\caption{Basic structure of transformer-based decoder model~\protect\cite{vaswani2017attention}. As shown in Figure~\ref{fig:tensors}, the parameters in the transformer part are managed using chunk-based memory, while the lm\_head part, responsible for deriving the output, consumes GPU memory based on its actual size.}
\label{fig:decoder}
\end{figure}

\subsection{Memory Consumption with Structure of Transformer-based Decoder Model}

The peak GPU memory usage on a single GPU (${m}_{peak}^{s}$) is
\begin{align*}
    m_{peak}^s=m_{base}+m_p+m_{os}+m_{out}+m_{lm}
\end{align*}
Each variable in this formula, except $m_{base}$ described in Section~\ref{section4-1}, is calculated as follows.

First, $m_p$ is the GPU memory used by param/gradient fp16 and param fp32. Considering that param fp16 and fp32 use 2 bytes ($B_{16}$) and 4 bytes ($B_{32}$) per value, respectively, $m_p$ is 
\begin{align*}
    m_p = \left\lceil (embed_p + \left\lceil \frac{other_p}{cs} \right\rceil \times cs) \times \frac{B_{16} + B_{32}}{cu_p}\right\rceil \times cu_p
\end{align*}
, where $embed_p$ is the input embedding param size, $cs$ is the chunk size, $other_p$ is the remaining param size, and $cu_p$ is the CUDA memory page size, typically $2 \times 1024^2$ bytes. Transformer-based decoder models~\cite{vaswani2017attention} are largely divided into a transformer model for fine-tuning and lm\_head for output, as shown in Figure~\ref{fig:decoder}. The part that uses the chunk memory is the transformer model in which the parameters are updated. The $embed_p$ is huge because the input embedding has a large dictionary. Therefore, $embed_p$ is managed separately.

Second, $m_{os}$ is the GPU memory used by momentum fp32 and variance fp32 of optimizer states. $m_{os}$ is 
\begin{align*}
    m_{os} = \sum_{t\in\{E,L\}}\left\lceil t_p\times\frac{B_{32}+B_{32}}{cu_p} \right\rceil \times cu_p
\end{align*}
, where $t$ is the operator of the given transformer model, $E$ is Embedding, $L$ is Linear, and $t_p$ is the parameter size of $t$. The system allocates GPU memory based on the actual size of each momentum fp32 and variance fp32, so GPU memory must be calculated for each tensor of each operator. Since the amount of GPU memory consumed by Bias or LayerNorm is very small, they can use space with other memory fragmentation. Therefore, we only calculate the GPU memory usage due to Embedding or Linear operator parameters.

Third, $m_{out}$ is the peak GPU memory usage due to output tensors. If 
the number of Embedding, layers, and model's output features are $e_n$, $l_n$, and $o_n$, respectively, then $m_{out}$ is 
\begin{align*}
    m_{out} = \left\lceil (e_n + l_n) \times (bs \times sl \times o_n) \times \frac{B_{16}}{cu_p} \right\rceil \times cu_p
\end{align*} 
PyTorch provides gradient checkpointing\footnote{Gradient checkpointing reduces GPU memory usage by clearing specific outputs and recomputing them during a backward pass.} as an option to save memory during fine-tuning. Therefore, we support estimating GPU memory usage due to each operator's input/output tensors considering gradient checkpointing. Since the output tensors of the current operator are the input tensors of the next operator, we focus on the output. 
It is challenging to accurately predict GPU memory consumption due to the outputs of operators within a model. We observed that the layer and embedding outputs of the transformer model are kept in GPU memory for efficient gradient checkpointing, which minimizes the increase in fine-tuning time. The estimation error rate is reduced using the $m_{out}$ equation, which accounts for our observation. 

\begin{figure}[t]
\begin{center}
    \includegraphics[width=0.99\linewidth]{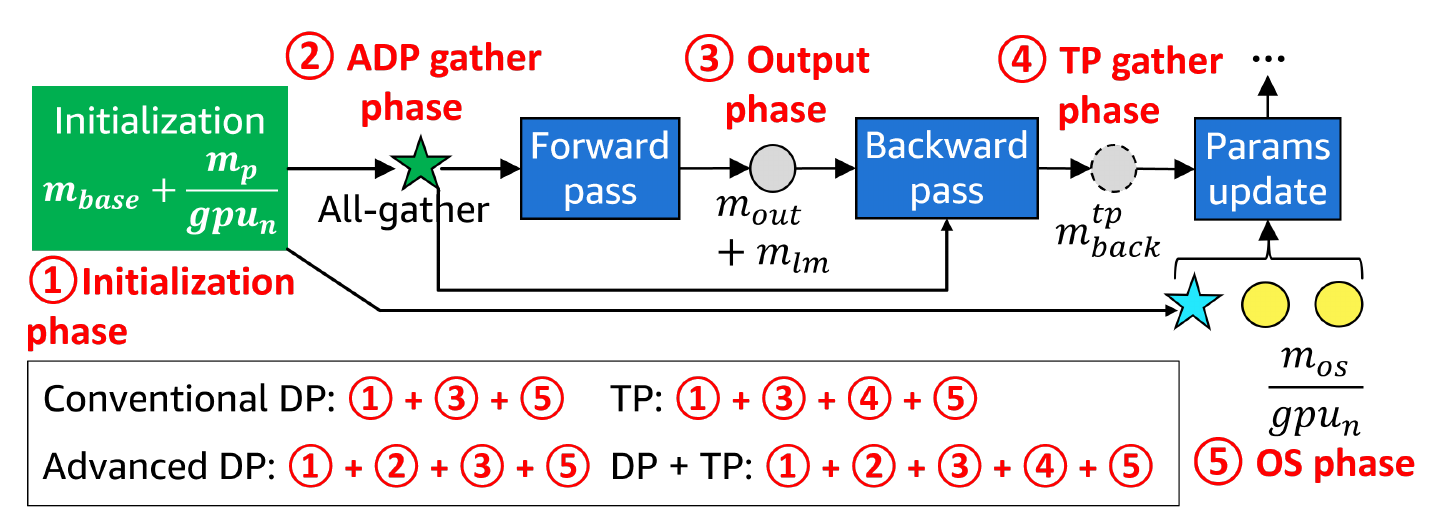}
\end{center}
\caption{Peak GPU memory computation for different distributed fine-tuning methods.}
\label{fig:peak-mems}
\end{figure}

Lastly, $m_{lm}$ is the GPU memory used in the lm\_head part including the loss calculation part. If the size of the embedding dictionary is $dict_n$, $m_{lm}$ is 
\begin{align*}
    m_{lm} = \left\lceil bs \times sl \times dict_n \times \frac{B}{cu_p} \right\rceil \times cu_p + \\
    2 \times \left\lceil bs \times (sl - 1) \times dict_n \times \frac{B}{cu_p} \right\rceil \times cu_p + lm_p
\end{align*}
, where $lm_p$ is the lm\_head param size, and $B$ is either $B_{16}$ or $B_{32}$ depending on the model type. The lm\_head converts the transformer outputs into logits. Then, the value obtained by shifting the sequence length of the logits by one space is stored in a separate temporary variable and used for the loss calculation. $m_{out} + m_{lm}$ is the output phase (\circled{3} in Figure~\ref{fig:peak-mems}).

\begin{figure}[t]
    \centering
    \begin{subfigure}[b]{0.31\textwidth}
        \centering
        \includegraphics[scale=0.25]{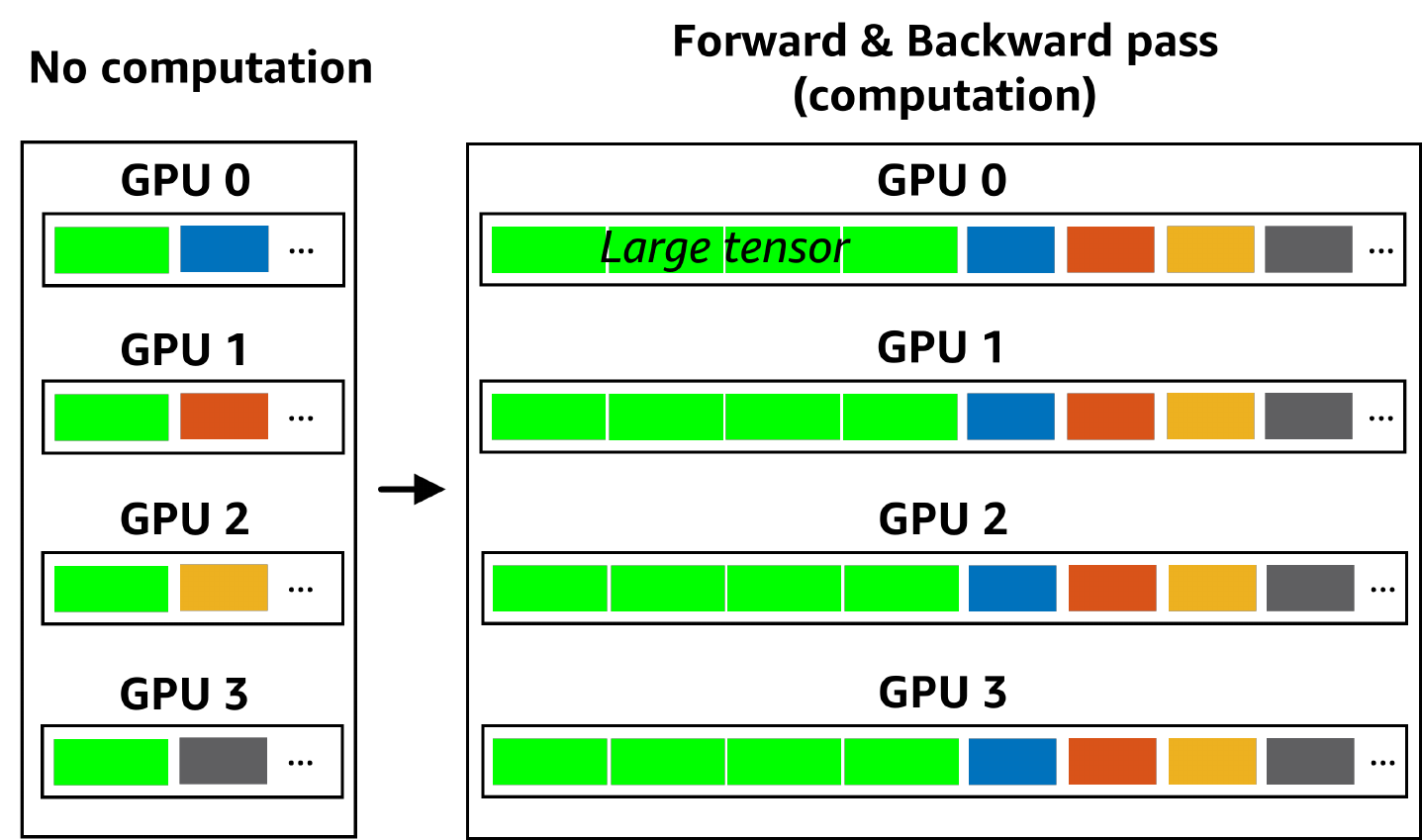}
        \caption{Advanced DP}
        \label{fig:dp-params}
    \end{subfigure}%
    ~ 
    \begin{subfigure}[b]{0.21\textwidth}
        \centering
        \includegraphics[scale=0.25]{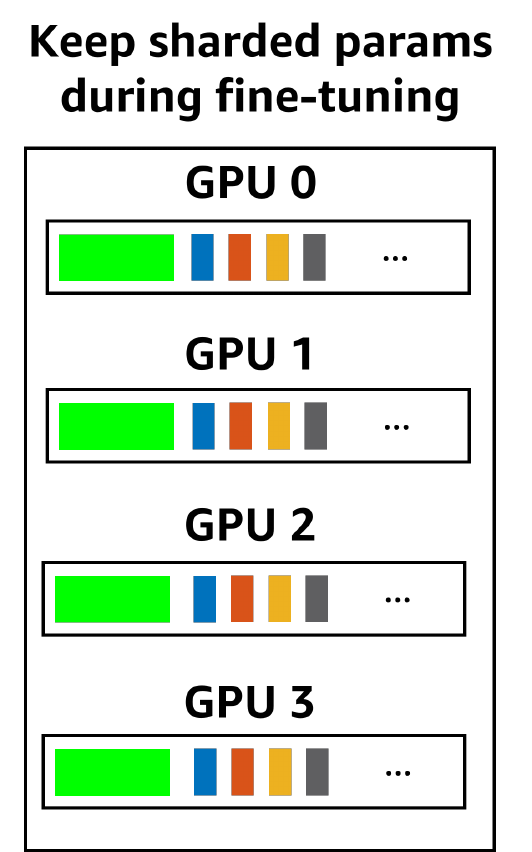}
        \caption{TP}
        \label{fig:tp-params}
    \end{subfigure}
    \caption{Advanced DP gathers the entire param fp16, while TP  maintains the sharded param fp16 intact before entering the computation process.}
\end{figure}

\begin{figure}[t]
\begin{center}
    \includegraphics[width=0.99\linewidth]{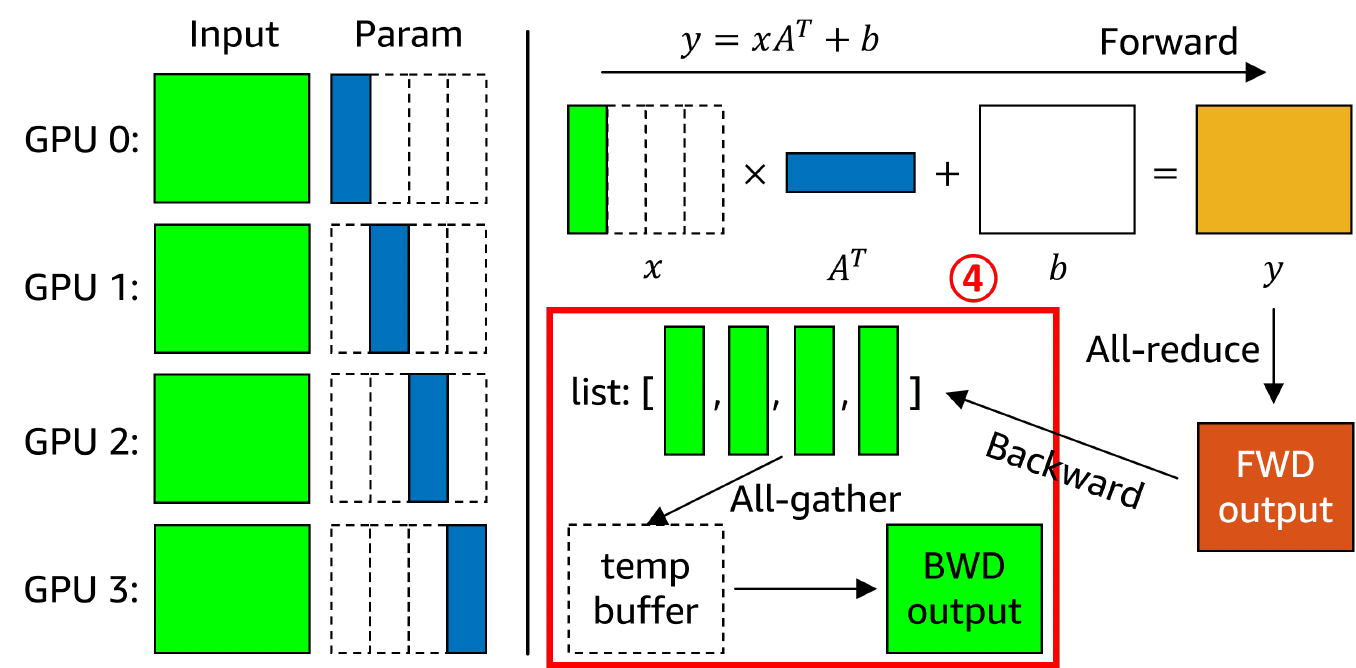}
\end{center}
\caption{Performing the linear operation in the forward and backward passes when employing TP. During the collection of partial outputs from each GPU after the backward pass, an additional GPU memory is consumed by a temporary buffer.}
\label{fig:tp-gather}
\end{figure}


\section{Multi-GPU Memory Usage Estimation} \label{section5}

This section outlines the factors for estimating peak GPU memory usage during distributed fine-tuning on multiple GPUs and summarizes the estimation process.

\noindent\textbf{Conventional data parallelism (CDP).} Since CDP places the entire model on each GPU, its peak GPU memory usage estimation equals the peak single-GPU memory usage $m_{peak}^s$, as shown in Figure~\ref{fig:peak-mems}.

\noindent\textbf{Advanced data parallelism (ADP).} The peak GPU memory usage with ADP on multiple GPUs ($m_{peak}^{dp}$) is 
\begin{align*}
    m_{peak}^{dp}=m_{base}+m_{p,16}+\frac{m_{p,32}+m_{os}}{gpu_n}+m_{out}+m_{lm}
\end{align*}
, where $m_{p,16}$ and $m_{p,32}$ are the GPU memory consumed by the entire param fp16/fp32, and $gpu_n$ is the number of GPUs in use. ZeRO-3 optimizer, a method of advanced data parallelism, evenly distributes parameters, gradients, and optimizer states by $gpu_n$, reducing GPU memory usage. Among these, gradient fp16 shares GPU memory with param fp16 as explained in Section~\ref{section4}, so we only need to divide the GPU memory usage of parameters and optimizer states by $gpu_n$. However, during the calculation process, each GPU must have all the values of param fp16 (Figure~\ref{fig:dp-params} and \circled{2} in Figure~\ref{fig:peak-mems}), so the whole $m_{p,16}$ is allocated to the GPU memory.

\noindent\textbf{Tensor parallelism (TP).} The peak GPU memory usage with 1D TP on multiple GPUs ($m_{peak}^{tp}$) is
\begin{align*}
    m_{peak}^{tp}=m_{base}+\frac{m_p+m_{os}}{gpu_n}+m_{out}+m_{lm}+m_{back}^{tp}
\end{align*}
, where $m_{back}^{tp}$ (\circled{4} in Figure~\ref{fig:peak-mems} and Figure~\ref{fig:tp-gather}) is the additional GPU memory usage due to the temporary buffer through the backward all-gather. If the number of GPUs used for tensor parallelism is $tp_n$, $m_{back}^{tp}$ is 
\begin{align*}
    m_{back}^{tp} = \left\lceil l_n \times (bs \times sl \times o_n) \times \frac{tp_n - 1}{tp_n} \times \frac{B_{16}}{cu_p} \right\rceil \times cu_p
\end{align*}
Tensor parallelism divides the parameter values of each operator by $gpu_n$ and does not combine them again, as shown in Figure~\ref{fig:tp-params}. It splits each model parameter tensor by row or column to apply tensor parallelism to multiple pre-trained language models. We call this one-dimension tensor parallelism (1D TP). Let us assume that we apply 1D TP to a linear operation on four GPUs. The linear operator's equation is $y=xA^T+b$, where $y$ is output, $x$ is input, $A$ is params/gradients, and $b$ is bias. The linear matrix multiplication process when each parameter tensor is split into columns is shown in Figure~\ref{fig:tp-gather}. We shard parameters by column because the output size after multiplication is the same as the size of the bias without sharding, so it is not affected by the use of bias. In the backward pass, the fine-tuning goes through an all-gather process. $m_{back}^{tp}$ is the total temporary buffer size for tensors imported from the other GPUs, calculated by multiplying the output size of each layer by the number of layers. 

\noindent\textbf{Combination of DP+TP.} The peak GPU memory usage with the combination of DP+TP on multiple GPUs ($m_{peak}^{dp+tp}$) is 
\begin{align*}
    m_{peak}^{dp+tp} = m_{peak}^{dp} - \frac{m_{p, 16} \times tp_n}{gpu_n} + m_{back}^{tp}
\end{align*}
, as shown in Figure~\ref{fig:peak-mems}. It is possible to achieve hybrid parallelism by fine-tuning through a combination of data and tensor parallelism. 
\section{Distributed Fine-Tuning Method Decision} \label{section6}

Algorithm~\ref{alg:dist_decision} describes the process for selecting the optimal method to fine-tune a pre-trained model based on the results of estimating the peak GPU memory usage. In Sections~\ref{section4} and~\ref{section5}, We estimated $m_{peak}^s$, $m_{peak}^{dp}$, $m_{peak}^{tp}$, and $m_{peak}^{dp+tp}$. Here, $m_{peak}^s$ represents CDP, 
and the remaining estimations are connected to ADP, TP, and DP+TP, respectively. Of these methods, the optimal one is the method that requires the shortest time for fine-tuning while avoiding GPU OOM.

\systemName\ takes a pre-trained model $M$, the total number of GPUs to fine-tune $gpu_n$, and the maximum sequence length $sl$. $eval$ is a list that stores the performance evaluation score of each method. $eval[0]$, $eval[1]$, $eval[2]$, and $eval[3]$ correspond to CDP, ADP, TP, and DP+TP, respectively. \systemName\ increments the batch size $bs$ for each method and gets the value of $bs$ when it reaches the total GPU memory capacity. Then, $bs - 1$ is the largest batch size to avoid GPU OOM. CDP uses $(bs - 1) \times gpu_n$ amount of data for fine-tuning in one iteration. In addition, since the ZeRO-3 optimizer increases the total communication volume of a baseline DP to $1.5\times$~\cite{rajbhandari2020zero}, the performance score of CDP is $(bs - 1) \times gpu_n \times 1.5$. In one iteration, ADP uses $(bs - 1) \times gpu_n$, TP uses $bs - 1$, and DP+TP uses $(bs - 1) \times dp_n$ of data for fine-tuning. $dp_n$ is the number of GPUs used for DP. These values become the performance scores of each method. Finally, \systemName\ selects the method with the highest performance score (If the scores are tied, select CDP, ADP, TP, and DP+TP in that order). If the performance scores of all methods are 0, heterogeneous training using CPU memory is selected as an alternative to avoid GPU OOM.

\begin{algorithm}[t]
\caption{Distributed Fine-Tuning Method Decision}\label{alg:dist_decision}
\small
\textbf{Input}: Pre-trained model $M$, $gpu_n$, and $sl$ \\
\textbf{Output}: Selected fine-tuning method and the optimal $bs$
\begin{algorithmic}[1]
    \State $eval = [0, 0, 0, 0]$ and $bs_{list} = [0, 0, 0, 0]$
    \State Measure total GPU memory capacity ($m_{total}$)
    \For{$i$ in range(4)}
        \State Set up the configure of the $i^{th}$ method
        \State $bs= 1$ and compute $m_{base}$, $m_p$, and $m_{os}$
        \State [1]: Compute $m_{out}$ and $m_{lm}$
        \If{$i = 0$}
            \State Repeat $bs = bs + 1$ until $m_{peak}^s > m_{total}$ after [1]
            \State $eval[i] = (bs - 1) \times gpu_n \times 1.5$
        \ElsIf{$i = 1$}
            \State Repeat $bs = bs + 1$ until $m_{peak}^{dp} > m_{total}$ after [1]
            \State $eval[i] = (bs - 1) \times gpu_n$
        \ElsIf{$i = 2$}
            \State Repeat $bs = bs + 1$ until $m_{peak}^{tp} > m_{total}$ after [1]
            \State $eval[i] = bs - 1$
        \ElsIf{$i = 3$}
            \State Repeat $bs = bs + 1$ until $m_{peak}^{dp+tp} > m_{total}$ after [1]
            \State $eval[i] = (bs - 1) \times dp_n$
        \EndIf
        \State $bs_{list}[i] = bs - 1$
    \EndFor
    \State Save the index of the maximum score to $idx$
    \State \Return $idx$, $bs_{list}[idx]$ \textbf{if} $eval[2] \neq 0$ \textbf{else} 4, 0
\end{algorithmic}
\end{algorithm}

\section{Experiments} \label{section7}

In this section, we compare the peak GPU memory usage estimate of \systemName\ with the ground truth data when applying various distributed fine-tuning methods. In addition, our DNNMem implementation is included in comparing GPU memory usage estimation on a single GPU.

\subsection{Experimental Setup} \label{setup}
For a multi-GPU environment, we use a Tesla V100 (total GPU memory capacity: 16384 MB) with 4 GPUs in CloudLab~\cite{cloudlab}. We also use the Colossal-AI~\cite{li2023colossal}, a widely used framework for applying distributed fine-tuning methods, and PyTorch 2.0.1 with CUDA 11.7. The models we used in the experiment are OPT~\cite{zhang2022opt}, BLOOM~\cite{workshop2022bloom}, CodeGen~\cite{nijkamp2022codegen}, BioGPT~\cite{luo2022biogpt}, GPTBigCode~\cite{allal2023santacoder}, GPT Neo~\cite{gpt-neo}, and LLaMA~\cite{touvron2023llama}. The dataset used is alpaca data~\cite{alpaca}, which is 52K instruction-following data. For the ground truth data, we measure peak GPU memory usage using only the maximum sequence length of 512.



\begin{figure}[t]
\begin{center}
    \includegraphics[width=0.9\linewidth]{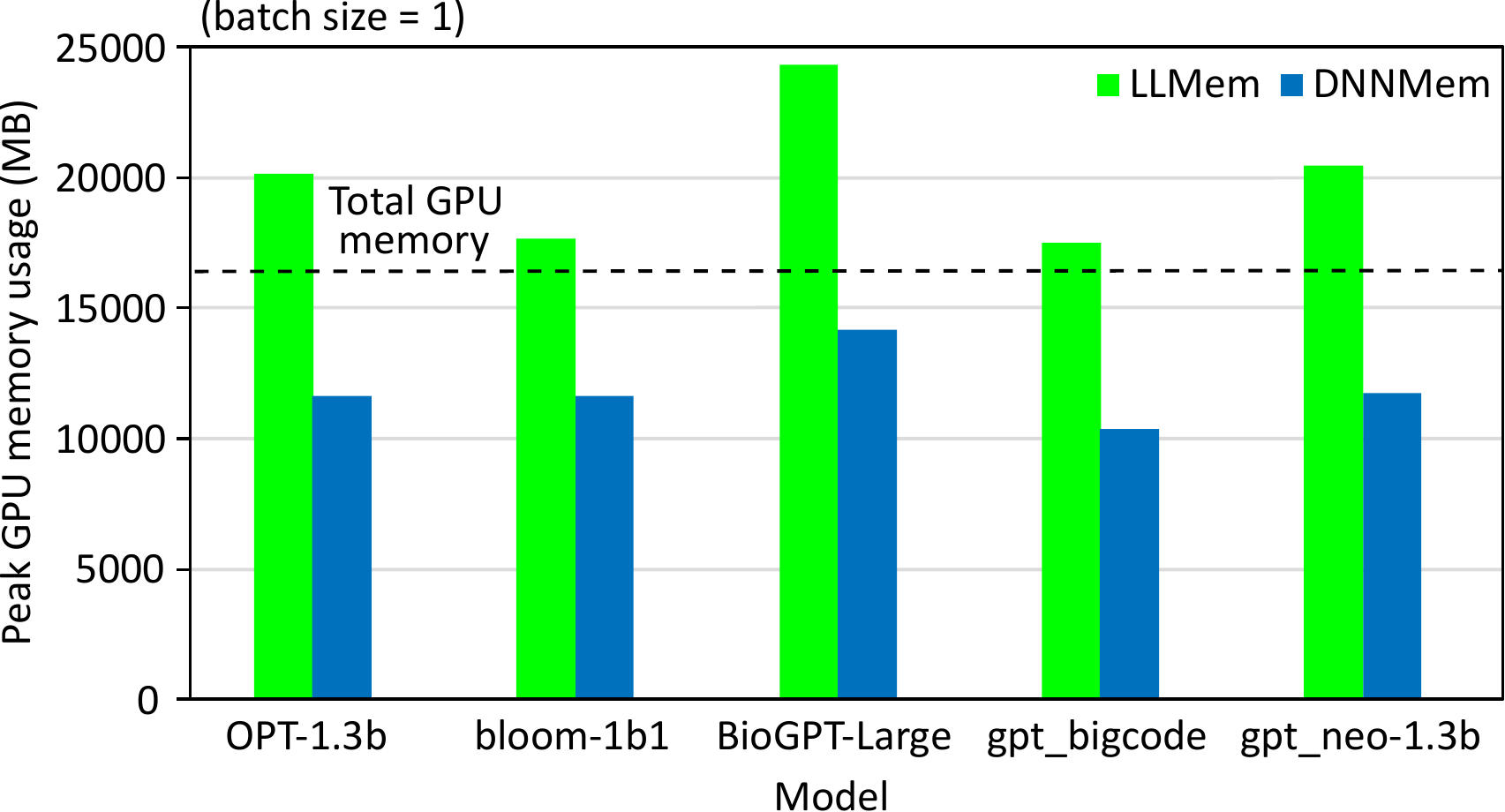}
\end{center}
\caption{Comparison of peak GPU memory usage estimates between \systemName\ and DNNMem for models experiencing GPU OOM during fine-tuning.}
\label{fig:oom-check}
\end{figure}

\begin{table}[t]
\caption{Estimating GPU memory usage on a single GPU. The values in parentheses represent the comparisons between the \systemName\ estimate and the ground truth, or the DNNMem estimate and the ground truth.}
\footnotesize 
\centering
\begin{tabular}{|l|l|l|l|}
\hline
Model (MB)   & \systemName\ & DNNMem & Ground truth  \\ \hline
OPT-125m     & 16314 (0.4)     & 10402 (36.5)  & 16378 \\
OPT-350m     & 16004 (1.6)     & 9354 (42.5)   & 16264 \\
bloom-560m   & 16578 (1.6)     & 10726 (34.3) & 16324 \\
codegen-350M & 16236 (0.8)     & 6910 (57.1)   & 16100 \\ \hline
\end{tabular}
\label{tab:single}
\end{table}

\subsection{Estimation of Single-GPU Memory Usage} \label{section7-2}

First, we compare the peak GPU memory usage estimate from \systemName\ for a single GPU with the DNNMem estimate and the actual peak GPU memory usage. Since we used gradient checkpointing for LLM fine-tuning, the same approach was applied to DNNMem. Figure~\ref{fig:oom-check} compares the peak GPU memory usage estimation results of \systemName\ and DNNMem for various pre-trained LLMs that cause GPU OOM during fine-tuning on a single GPU. \systemName\ predicts GPU OOM for all models, while DNNMem predicts peak GPU memory usage that falls short of $m_{total}$. Table~\ref{tab:single} shows the predicted and actual GPU memory usage peaks when applying the maximum batch size to obtain the ground truth data for each model during fine-tuning on a single GPU. DNNMem underestimates the peak GPU memory usage for all models because it does not account for factors considered when fine-tuning Transformer-based LLM, as explained in Section~\ref{section3-2}. \systemName's GPU memory estimation helps approximate the peak GPU memory usage close to the ground truth.

\begin{figure}[t]
\begin{center}
    \includegraphics[width=0.95\linewidth]{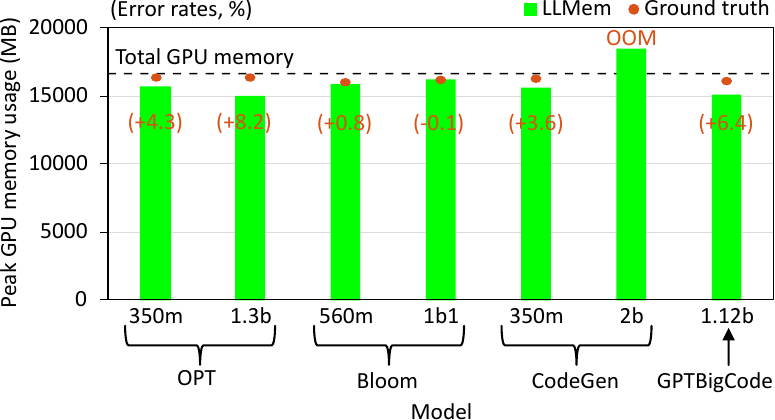}
\end{center}
\caption{Estimating GPU memory usage for ADP on four GPUs at each model's maximum batch size to prevent GPU OOM. OOM in codegen-2b indicates running out of memory even at batch size=1.}
\label{fig:dp}
\end{figure}

\subsection{Estimation of Multi-GPU Memory Usage}

\noindent\textbf{CDP.} The experimental results are the same as the memory usage estimation results on a single GPU in Section~\ref{section7-2}.

\noindent\textbf{ADP.} Figure~\ref{fig:dp} shows the predicted and actual GPU memory usage peaks when applying the maximum batch size to obtain ground truth data for each model during fine-tuning with ADP on four GPUs. The error rate between the predicted value of \systemName\ and the actual GPU memory usage tends to increase on multi-GPU setups. One reason is the gap in memory usage between the GPUs. ADP places tensors separately on each GPU instead of being sharded, so not all GPUs can use precisely the same amount of memory. Second, the error tends to be slightly larger when the model size is large. A larger number of layers and outputs in large models can lead to larger error rates due to memory allocator characteristics. 

\begin{figure}[t]
\begin{center}
    \includegraphics[width=0.99\linewidth]{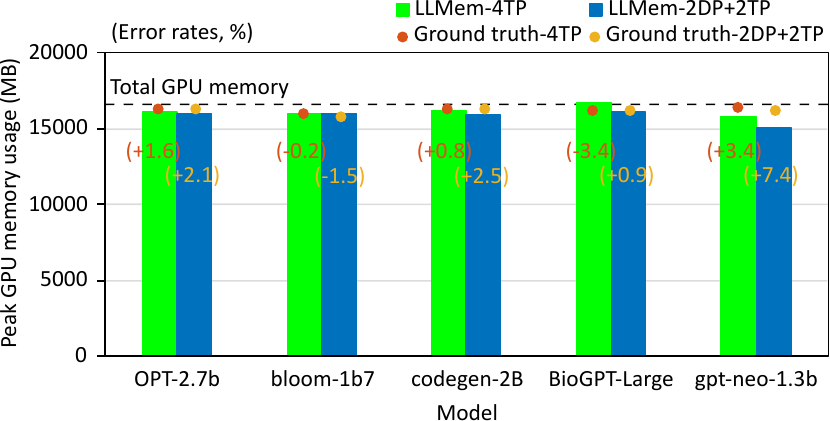}
\end{center}
\caption{Estimating GPU memory usage for 4TP and 2DP+2TP at each model's maximum batch size to avoid OOM.} 
\label{fig:tp}
\end{figure}

\noindent\textbf{TP and DP+TP.} Figure~\ref{fig:tp} shows the predicted and actual GPU memory usage peaks when applying the maximum batch size to obtain ground truth data for each model during fine-tuning with 4TP or 2DP+2TP on four GPUs. 4TP uses 4 GPUs in TP, and 2DP+2TP uses 2 GPUs in DP and 2 GPUs in TP for hybrid parallelism. 
We focus on estimating the peak GPU memory usage of the large-size model for TP because \systemName\ can select DP for quick fine-tuning of models that are small and do not have OOM problems.
TP applies the all-gather operation in the backward pass, as shown in Figure~\ref{fig:tp-gather}. The all-gather operation allocates temporary buffers in GPU memory and collects values in those buffers, consuming additional GPU memory. If the model size is large and the possible batch size is small, the system can use the allocated but currently empty memory space for a temporary buffer. Therefore, the GPU memory consumed due to the temporary buffer does not increase excessively, leading to smaller errors as shown in Figure~\ref{fig:tp}. 2DP+2TP shows slightly larger errors than 4TP in most cases. This is because GPU memory usage due to the temporary buffer may be additionally affected in \circled{2} and \circled{4} of Figure~\ref{fig:peak-mems} while applying both DP and TP.

\begin{table}[t]
\caption{Distributed fine-tuning method selection by \systemName\ on four GPUs and the actual amount of time it takes to fine-tune each method to the largest possible batch size (s)}
\scriptsize 
\centering
\begin{tabular}{|l|l|lll|}
\hline
Model (s)     & Selection   & 4DP           & 2DP+2TP & 4TP           \\ \hline
OPT-1.3b      & 4DP         & \textbf{688}  & 1616    & 2186          \\
OPT-2.7b      & 4TP         & OOM           & 8174    & \textbf{6038} \\
bloom-1b1     & 4DP         & \textbf{680}  & 1724    & 2631          \\
bloom-3b      & 4TP         & OOM           & OOM     & \textbf{14495}         \\
BioGPT-Large  & 4DP         & \textbf{1022} & 3315    & 4773          \\
codegen-2B-nl & 4TP         & OOM           & 6314    & \textbf{6244}          \\
gpt\_bigcode  & 4DP         & \textbf{651}  & 1292    & 1652          \\
gpt-neo-1.3B  & 4DP         & \textbf{768}  & 1686    & 2372          \\
llama-7b      & CPU offloading & OOM           & OOM     & OOM           \\ \hline
\end{tabular}
\label{tab:selection}
\end{table}

\subsection{Fine-Tuning Method Selection with \systemName}

Table~\ref{tab:selection} assesses whether \systemName\ finds the optimal 
fine-tuning method to achieve the fastest fine-tuning while avoiding GPU OOM for various models. When measuring the time taken for each method, we applied the maximum batch size that can prevent GPU OOM. \systemName\ typically selects TP when DP causes GPU OOM. It is challenging for \systemName\ to choose DP+TP because only 4 GPUs were used in the experiment. DP+TP allows for more diverse combinations depending on the number of GPUs used and is more likely to be selected. \systemName\ also suggests CPU offloading when GPU memory is insufficient. 
\section{Conclusion} \label{section8}

This paper introduces \systemName, a method for estimating GPU memory consumption during fine-tuning of large language models (LLMs) on multi-GPU setups. We analyze factors affecting GPU memory usage, considering different memory allocation methods for the transformer and output sections. Experimental results demonstrate that \systemName\ achieves accurate peak GPU memory usage estimation on both single and multiple GPUs with minimal error rates.

\bibliographystyle{named}
\bibliography{ijcai24}

\begin{thebibliography}{}

\bibitem[\protect\citeauthoryear{Albahar \bgroup \em et al.\egroup }{2022}]{albahar2022schedtune}
Hadeel Albahar, Shruti Dongare, Yanlin Du, Nannan Zhao, Arnab~K Paul, and Ali~R Butt.
\newblock Schedtune: A heterogeneity-aware gpu scheduler for deep learning.
\newblock In {\em 2022 22nd IEEE International Symposium on Cluster, Cloud and Internet Computing (CCGrid)}, pages 695--705. IEEE, 2022.

\bibitem[\protect\citeauthoryear{Allal \bgroup \em et al.\egroup }{2023}]{allal2023santacoder}
Loubna~Ben Allal, Raymond Li, Denis Kocetkov, Chenghao Mou, Christopher Akiki, Carlos~Munoz Ferrandis, Niklas Muennighoff, Mayank Mishra, Alex Gu, Manan Dey, et~al.
\newblock Santacoder: don't reach for the stars!
\newblock {\em arXiv preprint arXiv:2301.03988}, 2023.

\bibitem[\protect\citeauthoryear{Bian \bgroup \em et al.\egroup }{2021}]{bian2021maximizing}
Zhengda Bian, Qifan Xu, Boxiang Wang, and Yang You.
\newblock Maximizing parallelism in distributed training for huge neural networks.
\newblock {\em arXiv preprint arXiv:2105.14450}, 2021.

\bibitem[\protect\citeauthoryear{Black \bgroup \em et al.\egroup }{2021}]{gpt-neo}
Sid Black, Gao Leo, Phil Wang, Connor Leahy, and Stella Biderman.
\newblock Gpt-neo: Large scale autoregressive language modeling with mesh-tensorflow.
\newblock \url{https://doi.org/10.5281/zenodo.5297715}, 2021.

\bibitem[\protect\citeauthoryear{CloudLab}{2023}]{cloudlab}
CloudLab.
\newblock \url{https://www.cloudlab.us/}, 2023.

\bibitem[\protect\citeauthoryear{Devlin \bgroup \em et al.\egroup }{2018}]{devlin2018bert}
Jacob Devlin, Ming-Wei Chang, Kenton Lee, and Kristina Toutanova.
\newblock Bert: Pre-training of deep bidirectional transformers for language understanding.
\newblock {\em arXiv preprint arXiv:1810.04805}, 2018.

\bibitem[\protect\citeauthoryear{Fang \bgroup \em et al.\egroup }{2022}]{fang2022parallel}
Jiarui Fang, Zilin Zhu, Shenggui Li, Hui Su, Yang Yu, Jie Zhou, and Yang You.
\newblock Parallel training of pre-trained models via chunk-based dynamic memory management.
\newblock {\em IEEE Transactions on Parallel and Distributed Systems}, 34(1):304--315, 2022.

\bibitem[\protect\citeauthoryear{Gao \bgroup \em et al.\egroup }{2020}]{gao2020estimating}
Yanjie Gao, Yu~Liu, Hongyu Zhang, Zhengxian Li, Yonghao Zhu, Haoxiang Lin, and Mao Yang.
\newblock Estimating gpu memory consumption of deep learning models.
\newblock In {\em Proceedings of the 28th ACM Joint Meeting on European Software Engineering Conference and Symposium on the Foundations of Software Engineering}, pages 1342--1352, 2020.

\bibitem[\protect\citeauthoryear{Li \bgroup \em et al.\egroup }{2023}]{li2023colossal}
Shenggui Li, Hongxin Liu, Zhengda Bian, Jiarui Fang, Haichen Huang, Yuliang Liu, Boxiang Wang, and Yang You.
\newblock Colossal-ai: A unified deep learning system for large-scale parallel training.
\newblock In {\em Proceedings of the 52nd International Conference on Parallel Processing}, pages 766--775, 2023.

\bibitem[\protect\citeauthoryear{Luo \bgroup \em et al.\egroup }{2022}]{luo2022biogpt}
Renqian Luo, Liai Sun, Yingce Xia, Tao Qin, Sheng Zhang, Hoifung Poon, and Tie-Yan Liu.
\newblock Biogpt: generative pre-trained transformer for biomedical text generation and mining.
\newblock {\em Briefings in Bioinformatics}, 23(6):bbac409, 2022.

\bibitem[\protect\citeauthoryear{Nie \bgroup \em et al.\egroup }{2022}]{nie2022tsplit}
Xiaonan Nie, Xupeng Miao, Zhi Yang, and Bin Cui.
\newblock Tsplit: Fine-grained gpu memory management for efficient dnn training via tensor splitting.
\newblock In {\em 2022 IEEE 38th International Conference on Data Engineering (ICDE)}, pages 2615--2628. IEEE, 2022.

\bibitem[\protect\citeauthoryear{Nijkamp \bgroup \em et al.\egroup }{2022}]{nijkamp2022codegen}
Erik Nijkamp, Bo~Pang, Hiroaki Hayashi, Lifu Tu, Huan Wang, Yingbo Zhou, Silvio Savarese, and Caiming Xiong.
\newblock Codegen: An open large language model for code with multi-turn program synthesis.
\newblock {\em arXiv preprint arXiv:2203.13474}, 2022.

\bibitem[\protect\citeauthoryear{Rajbhandari \bgroup \em et al.\egroup }{2020}]{rajbhandari2020zero}
Samyam Rajbhandari, Jeff Rasley, Olatunji Ruwase, and Yuxiong He.
\newblock Zero: Memory optimizations toward training trillion parameter models.
\newblock In {\em SC20: International Conference for High Performance Computing, Networking, Storage and Analysis}, pages 1--16. IEEE, 2020.

\bibitem[\protect\citeauthoryear{Ren \bgroup \em et al.\egroup }{2021}]{ren2021zero}
Jie Ren, Samyam Rajbhandari, Reza~Yazdani Aminabadi, Olatunji Ruwase, Shuangyan Yang, Minjia Zhang, Dong Li, and Yuxiong He.
\newblock $\{$ZeRO-Offload$\}$: Democratizing $\{$Billion-Scale$\}$ model training.
\newblock In {\em 2021 USENIX Annual Technical Conference (USENIX ATC 21)}, pages 551--564, 2021.

\bibitem[\protect\citeauthoryear{Shoeybi \bgroup \em et al.\egroup }{2019}]{shoeybi2019megatron}
Mohammad Shoeybi, Mostofa Patwary, Raul Puri, Patrick LeGresley, Jared Casper, and Bryan Catanzaro.
\newblock Megatron-lm: Training multi-billion parameter language models using model parallelism.
\newblock {\em arXiv preprint arXiv:1909.08053}, 2019.

\bibitem[\protect\citeauthoryear{Taori \bgroup \em et al.\egroup }{2023}]{alpaca}
Rohan Taori, Ishaan Gulrajani, Tianyi Zhang, Yann Dubois, Xuechen Li, Carlos Guestrin, Percy Liang, and Tatsunori~B. Hashimoto.
\newblock Stanford alpaca: An instruction-following llama model.
\newblock \url{https://github.com/tatsu-lab/stanford_alpaca}, 2023.

\bibitem[\protect\citeauthoryear{Touvron \bgroup \em et al.\egroup }{2023}]{touvron2023llama}
Hugo Touvron, Thibaut Lavril, Gautier Izacard, Xavier Martinet, Marie-Anne Lachaux, Timoth{\'e}e Lacroix, Baptiste Rozi{\`e}re, Naman Goyal, Eric Hambro, Faisal Azhar, et~al.
\newblock Llama: Open and efficient foundation language models.
\newblock {\em arXiv preprint arXiv:2302.13971}, 2023.

\bibitem[\protect\citeauthoryear{Vaswani \bgroup \em et al.\egroup }{2017}]{vaswani2017attention}
Ashish Vaswani, Noam Shazeer, Niki Parmar, Jakob Uszkoreit, Llion Jones, Aidan~N Gomez, {\L}ukasz Kaiser, and Illia Polosukhin.
\newblock Attention is all you need.
\newblock {\em Advances in neural information processing systems}, 30, 2017.

\bibitem[\protect\citeauthoryear{Wang \bgroup \em et al.\egroup }{2018}]{wang2018glue}
Alex Wang, Amanpreet Singh, Julian Michael, Felix Hill, Omer Levy, and Samuel~R Bowman.
\newblock Glue: A multi-task benchmark and analysis platform for natural language understanding.
\newblock {\em arXiv preprint arXiv:1804.07461}, 2018.

\bibitem[\protect\citeauthoryear{Wang \bgroup \em et al.\egroup }{2021}]{wang20212}
Boxiang Wang, Qifan Xu, Zhengda Bian, and Yang You.
\newblock 2.5-dimensional distributed model training.
\newblock {\em arXiv e-prints}, pages arXiv--2105, 2021.

\bibitem[\protect\citeauthoryear{Workshop \bgroup \em et al.\egroup }{2022}]{workshop2022bloom}
BigScience Workshop, Teven~Le Scao, Angela Fan, Christopher Akiki, Ellie Pavlick, Suzana Ili{\'c}, Daniel Hesslow, Roman Castagn{\'e}, Alexandra~Sasha Luccioni, Fran{\c{c}}ois Yvon, et~al.
\newblock Bloom: A 176b-parameter open-access multilingual language model.
\newblock {\em arXiv preprint arXiv:2211.05100}, 2022.

\bibitem[\protect\citeauthoryear{Xu \bgroup \em et al.\egroup }{2021}]{xu2021efficient}
Qifan Xu, Shenggui Li, Chaoyu Gong, and Yang You.
\newblock An efficient 2d method for training super-large deep learning models.
\newblock {\em arXiv e-prints}, pages arXiv--2104, 2021.

\bibitem[\protect\citeauthoryear{Zhang \bgroup \em et al.\egroup }{2022}]{zhang2022opt}
Susan Zhang, Stephen Roller, Naman Goyal, Mikel Artetxe, Moya Chen, Shuohui Chen, Christopher Dewan, Mona Diab, Xian Li, Xi~Victoria Lin, et~al.
\newblock Opt: Open pre-trained transformer language models.
\newblock {\em arXiv preprint arXiv:2205.01068}, 2022.

\end{thebibliography}

\end{document}